\documentclass[letterpaper, 9pt, conference]{ieeeconf}  % Comment this line out if you need a4paper

\IEEEoverridecommandlockouts   
\usepackage{cite}
\usepackage{amsmath,amssymb,amsfonts}
\usepackage{tabularx,booktabs}
\usepackage{algorithmic}
\usepackage{graphicx}
\usepackage{xcolor}
\usepackage{textcomp}

\title{GBEC: Geometry-Based Hand-Eye Calibration}
\author{Yihao Liu, Jiaming Zhang, Zhangcong She, Amir Kheradmand$^\dagger$, Mehran Armand$^\dagger$ 
\thanks{Manuscript submitted Sept 15, 2023. This work was supported by grants from the National Institute of Deafness and Other Communication Disorders (R01DC018815), National Institute of Arthritis and Musculoskeletal and Skin Diseases (R01AR080315), and National Institute of Biomedical Imaging and Bioengineering (R01EB023939).} 
\thanks{Yihao Liu is with the Department of Computer Science, Johns Hopkins University, Baltimore, MD 21218 USA (Corresponding author e-mail: yliu333@jhu.edu). }
\thanks{Jiaming Zhang is with the Department of Computer Science, Johns Hopkins University, Baltimore, MD 21218 USA. }
\thanks{Zhangcong She is with the Department of Mechanical Engineering, Johns Hopkins University, Baltimore, MD 21218 USA. }
\thanks{Amir Kheradmand is with the Department of Neurology, and Department of Neuroscience, Johns Hopkins University, Baltimore, MD 21218 USA.}
\thanks{Mehran Armand is with the Department of Mechanical Engineering, Department of Computer Science, Department of Orthopaedic Surgery, and Applied Physics Laboratory, Johns Hopkins University, Baltimore, MD 21218 USA.}
\thanks{$\dagger$ Equal contribution}
}

\begin{document}

\maketitle

\begin{abstract}
Hand-eye calibration is the problem of solving the transformation from the end-effector of a robot to the sensor attached to it. Commonly employed techniques, such as AXXB or AXZB formulations, rely on regression methods that require collecting pose data from different robot configurations, which can produce low accuracy and repeatability. However, the derived transformation should solely depend on the geometry of the end-effector and the sensor attachment. We propose Geometry-Based End-Effector Calibration (GBEC) that enhances the repeatability and accuracy of the derived transformation compared to traditional hand-eye calibrations. To demonstrate improvements, we apply the approach to two different robot-assisted procedures: Transcranial Magnetic Stimulation (TMS) and femoroplasty. We also discuss the generalizability of GBEC for camera-in-hand and marker-in-hand sensor mounting methods. In the experiments, we perform GBEC between the robot end-effector and an optical tracker's rigid body marker attached to the TMS coil or femoroplasty drill guide. Previous research documents low repeatability and accuracy of the conventional methods for robot-assisted TMS hand-eye calibration. Applying GBEC to repeated calibrations, we obtain transformations with standard deviations of 0.37mm, 0.65mm, and 0.40mm (translation) along x, y, and z axes of the end-effector, respectively. The tool alignment experiments after using GBEC achieve a mean accuracy around 0.2mm in Euclidean distance. When compared to some existing methods, the proposed method relies solely on the geometry of the flange and the pose of the rigid-body marker, making it independent of workspace constraints or robot accuracy, without sacrificing the orthogonality of the rotation matrix. Our results validate the accuracy and applicability of the approach, providing a new and generalizable methodology for obtaining the transformation from the end-effector to a sensor. 
\end{abstract}

\begin{keywords}
Hand-Eye Calibration, Robot-Assisted Transcranial Magnetic Stimulation, Robot-Assisted Femoroplasty
\end{keywords}

\section{Introduction}
\label{sec:introduction}

Finding the transformation between a robot end-effector and its sensory attachments, known as hand-eye calibration, is essential for accurate planning and control of the robot motion. Solving the hand-eye calibration problem can be affected by factors such as sensor noise, robot range of motion, inaccurate kinematic models, or optimization algorithm \cite{ackerman2013probabilistic, condurache2019novel, heller2014hand, hsiao2020positioning, li2015simultaneous, ma2016new, lu2023error, brookshire2016articulated, enebuse2021comparative, enebuse2022accuracy, gwak2003numerical, zhang2017computationally, zhao2019simultaneous, zhong2020hand, zhou2018towards}. In vision-based methods, the accuracy can also be affected by image processing, ambient conditions, and calibration target errors. In medical robots, the inaccuracy of hand-eye calibration can affect the alignment of the tool attached to the robot with the anatomy of interest \cite{zhang2017computationally, zhong2020hand, zhou2018towards, richter2011robust, ernst2012non, richter2013optimal, noccaro2021development}. The problem is especially prominent in robot-assisted TMS, our main motivating clinical application.

\begin{figure}[t]
    \centering
    \includegraphics[width=0.38\textwidth]{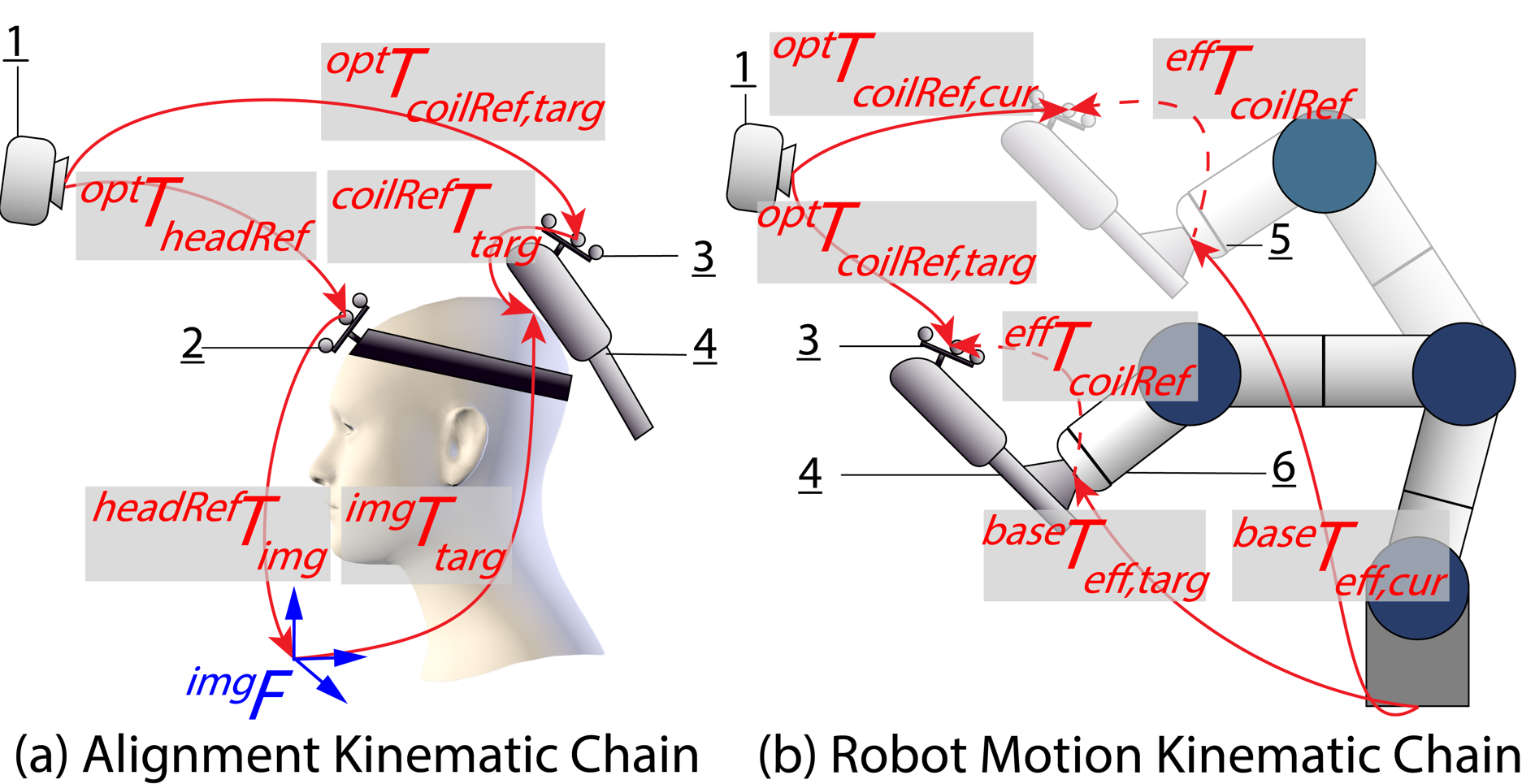}
    \caption{Kinematic chains of robot-assisted TMS. The numbers in Fig. 1(a) and 1(b) refer to: \underline{1} the optical tracking camera, \underline{2} the RBM affixed to the subject's head, enabling the acquisition of spatial information, \underline{3} the RBM affixed on the TMS coil, \underline{4} the TMS coil, \underline{5} the robotic arm (depicted in semi-transparent) at its initial pose, and \underline{6} the robotic arm at the aligned pose. The dashed line arrows labeled $^{eff}T_{coilRef}$ represent the transformation between the end-effector and the RBM affixed to the TMS coil. This transformation is the derived result of the proposed GBEC.}
    \label{fig:alignkinematics}
\end{figure}

TMS is a non-invasive technique for brain modulation by applying magnetic fields to stimulate nerve cells in the brain \cite{hallett2007transcranial}. It has diverse clinical and research applications, including neuropsychiatry \cite{george1999transcranial}, neurology \cite{kobayashi2003transcranial}, and cognitive neuroscience \cite{walsh2000transcranial}. Commonly, TMS coils are positioned on the subject's head by human operators, relying on neuronavigation systems for manual alignment \cite{sack2009optimizing}. The accuracy of manual coil placement largely depends on the operator, with an average error ranging 5-6mm \cite{sack2009optimizing, lefaucheur2010image, schoenfeldt2010value}. Robot-assisted TMS can address this limitation by improving the accuracy of TMS coil placement on the head \cite{zorn2011design, goetz2019accuracy, kantelhardt2010robot}. The kinematic chain associated with this task is depicted in Fig. \ref{fig:alignkinematics}. Robot-assisted TMS, if using an optical tracking camera and an rigid body marker (RBM) on the TMS coil, requires hand-eye calibration to establish the relationship between the end-effector of the robot to the RBM \cite{richter2011robust, ernst2012non, richter2013optimal, noccaro2021development}. Existing calibration methods \cite{tsai1989new, shah2012overview, strobl2006optimal}, including those specific to robot-assisted TMS, suffer from limitations such as low repeatability, restricted workspace, sacrificed orthogonality of the rotation matrix, zero motion of the robot base with respect to the fixation platform, and reliance on robot kinematics accuracy. 

In this study, we propose a novel approach to hand-eye calibration, leveraging the geometry of the end-effector. Compared to conventional AXXB and AXZB formulations, the approach enables repeatable and accurate outcomes, eliminating the reliance on the workspace and robot accuracy. We apply this approach to the robot-assisted TMS, and analyze and discuss the sources of errors in detail. Furthermore, we extend GBEC to different landmark configuration (robot-assisted femoroplasty) and different sensor mounting methods (camera-in-hand), and present the generalization conditions.

\begin{figure}[t]
    \centering
    \includegraphics[width=0.5\textwidth]{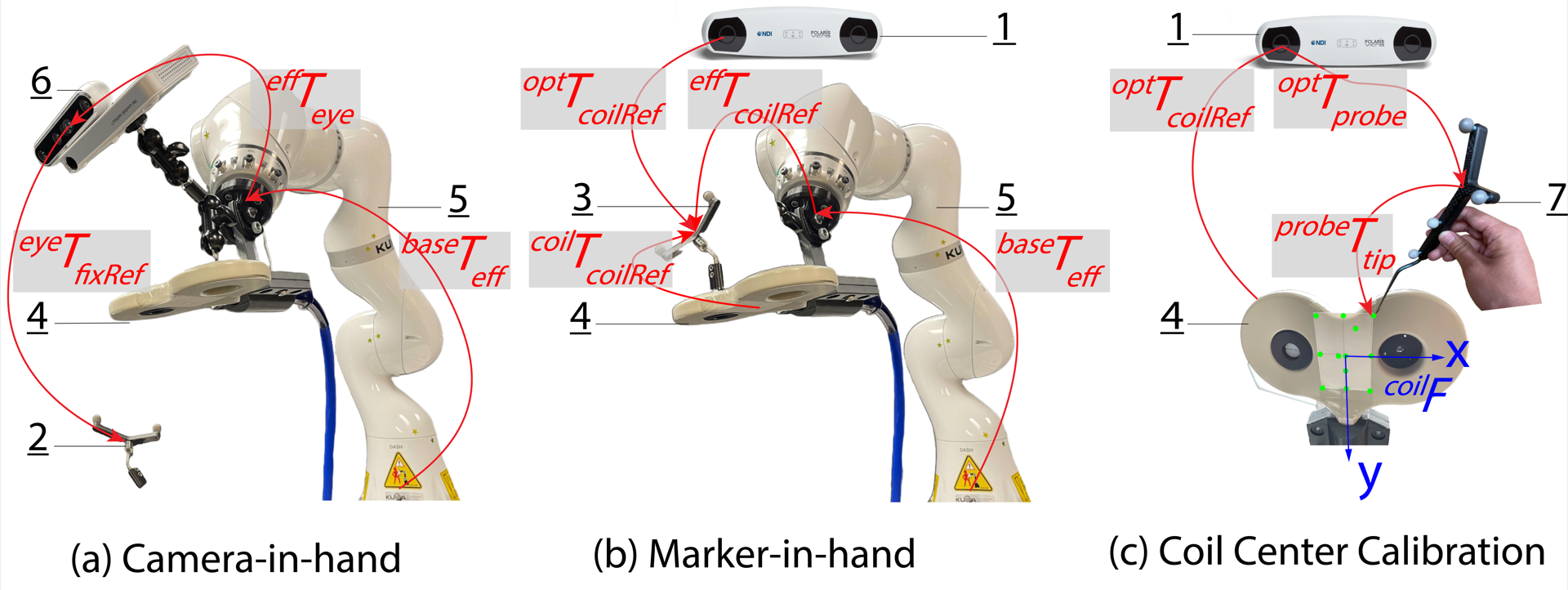}
    \caption{The ``camera-in-hand'' and ``marker-in-hand'' setups. The numbers \underline{1}-\underline{5} match the numbering in Fig. \ref{fig:alignkinematics}. Here, \underline{6} is the Portable Projection Mapping Device used in \cite{liu2022inside}, and \underline{7} is a rigid body probe used to digitize the position of a point.}
    \label{fig:kinematics}
\end{figure}

\begin{figure}[t]
    \centering
    \includegraphics[width=0.4\textwidth]{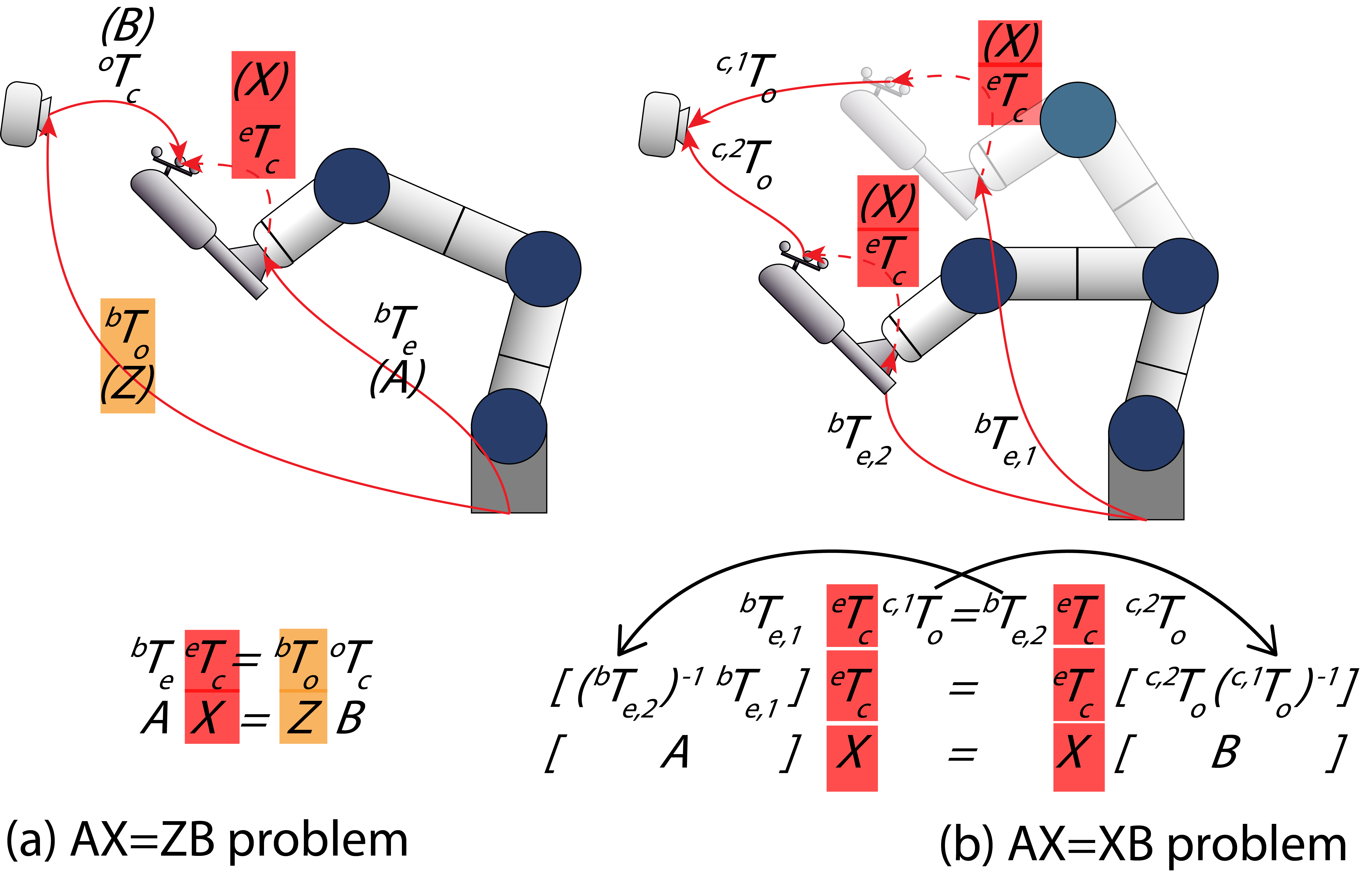}
    \caption{The formulation of $AX=ZB$ and $AX=XB$ problems. To simplify the equations, here, the superscripts and subscripts from Fig. \ref{fig:alignkinematics} are simplified to the initial letters. In this figure, unknown transformations are in red and orange backgrounds. In (b), the black arrows on the equations indicate rearrangement of the transformations to formulate $AX=XB$.}
    \label{fig:handeye}
\end{figure}

\section{Related Work}\label{sec:previouswork}

Calibrations for TMS coil are mainly to derive two transformations, one between the center of the coil and the coil RBM ($^{coil}T_{coilRef}$, coil center calibration) and the other between the robot arm end-effector and the coil RBM ($^{eff}T_{coilRef}$, hand-eye calibration). The transformations are illustrated in Fig. \ref{fig:alignkinematics}, and the physical setup for marker-in-hand (outside-in tracking \footnote{The terms ``inside-out'' and ``outside-in'' in \cite{liu2022inside} are used under the context of tracking methods in mixed reality applications. In the content of calibration, ``camera-in-hand'' and ``marker-in-hand'' are used, corresponding to the calibration setups in inside-out tracking and outside-in tracking, respectively.}) is shown in Fig. \ref{fig:kinematics}(b). The work in \cite{liu2022inside} shows a paired-point registration method for coil-center calibration ($^{coil}T_{coilRef}$) and an $AX=XB$ algorithm for hand-eye calibration ($^{eff}T_{coilRef}$). For coil-center calibration, we use a sticker attached to the coil-center. The sticker contains pre-defined landmarks measured with respect to a reference frame established at the center of the coil, as shown in Fig. \ref{fig:kinematics}(c). This geometry-based coil-center calibration method has been reported to produce repeatable and submillimeter error residuals \cite{liu2022inside}. On the other hand, the hand-eye calibration performed using $AX=XB$, has shown low repeatability. GBEC improves hand-eye calibration results by adopting the same accurate geometry-based methodology in the TMS coil center calibration method in \cite{liu2022inside} instead of using $AX=XB$. 

Shown in Fig. \ref{fig:handeye}, hand-eye calibration can be formulated as an $AX=XB$ or $AX=ZB$ problem \cite{tabb2017solving, shah2012overview}. Both $AX=ZB$ and $AX=XB$ have been studied extensively in the literature and have mainly three types of solutions: Separable, simultaneous and iterative \cite{shah2012overview, tabb2017solving}. Most hand-eye calibration solutions are regression algorithms that use data $A,B$ collected at different robot joint configurations. However, in theory, the unknown transformation ($X$) should only depend on the position of the RBM on the end-effector and the definition of the center of the end-effector, which are both geometrically independent of the pose of the robot. Consequently, the data obtained from the robot can introduce errors in the calibration result. Previous studies approach hand-eye calibration in robot-assisted TMS using $AX=XB$ or $AX=ZB$ methods \cite{richter2011robust, richter2013optimal, ernst2012non, noccaro2021development}. The methods are mainly (1) using a non-orthogonal rotation matrix, (2) limiting the workspace of the calibration, or (3) updating calibration online using the real-time state of the robot joints.

Ernst \textit{et al}. proposed a non-orthogonal calibration method that compensates for calibration errors by allowing for non-orthogonal matrices \cite{ernst2012non}. The study highlights that the data from the robots are not accurate, and their errors can range from 2-3 mm. The optimization process in solving hand-eye calibration relies on the collected data from different robot configurations, therefore, adding to the error. By using non-orthogonal matrices for calibration and allowing additional degrees of freedom in matrices, the calibration procedure can correct system inaccuracies that rigid body transformations fail to account for. However, despite producing a non-orthogonal matrix that addresses system inaccuracies, Ernst \textit{et al.} found it necessary to apply orthonormalization to their final calibration result as most applications require orthogonal matrices \cite{ernst2012non}. The implications of orthonormalization on the resulting calibration accuracy are not clear. Noccaro \textit{et al}. proposed to constrain the calibration method by the specific application \cite{noccaro2021development}. For robot-assisted TMS, they limited the data collection for calibration to the TMS workspace around the head of the subject. The study showed empirical improvement. However, restricting the workspace may constrain the motion of the payload and the center of gravity of the robot to a small space, neglecting the potential deformation in the fixation platform where the robot base is mounted. Thus, the collected data is more accurate at the limited space, since $AX=XB$ and $AX=ZB$ problems assume that the robot base remains fixed while moving to different poses. Richter \textit{et al.} \cite{richter2011robust} finds low repeatability in non-orthogonal QR24-algorithm \cite{ernst2012non} and the standard AX=XB algorithm \cite{tsai1989new}, comparing them to their own proposed online calibration algorithm. The findings revealed significant variation in the translation components of the transformation matrix ranging up to 10mm.

Most methods reported in the literature solve $AX=XB$ or $AX=ZB$ problems by modifying data collection and regression process. These methods treat the robot as a sensor, wherein $B$ represents or contains end-effector poses, and $A$ represents or contains optical tracking camera data. Optical tracking cameras are typically considered to be accurate, and the system (Polaris Vicra, NDI) used in our work has errors of around $0.25mm$ \cite{koivukangas2013technical}. However, it is challenging to achieve the same level of accuracy with most robots. Errors can arise not only from the robot's actuation but also from suboptimal fixation. 

The hand-eye calibration result should only depend on the geometry of the flange and the location of the RBM. The reason that the previous methods depend on the collected data from the robot is that they treat the robot as a sensor. The fundamental concept of our work is to only trust reliable optical tracking cameras for data collection and use geometry to infer the transformation between the end-effector and the marker. By eliminating data collection based on robot poses, the calibration method becomes independent of the robot poses and workspace considerations. 

\section{Method}\label{sec:methods}

This section introduces the implementation details of GBEC including the core algorithm, the application on the end-effector used in robot-assisted TMS, fiducial (landmark) localization error and its optimization, and the versatility of GBEC in another medical application (robot-assisted femoroplasty). Notebly, hand-eye calibration is used for two different sensor mounting modes: Camera-in-hand and marker-in-hand. Robot-assisted TMS and robot-assisted femoroplasty use marker-in-hand mode. The extension of camera-in-hand to GBEC is discussed in Section \ref{sec:future}. 

\begin{figure}[t]
    \centering
    \includegraphics[width=0.35\textwidth]{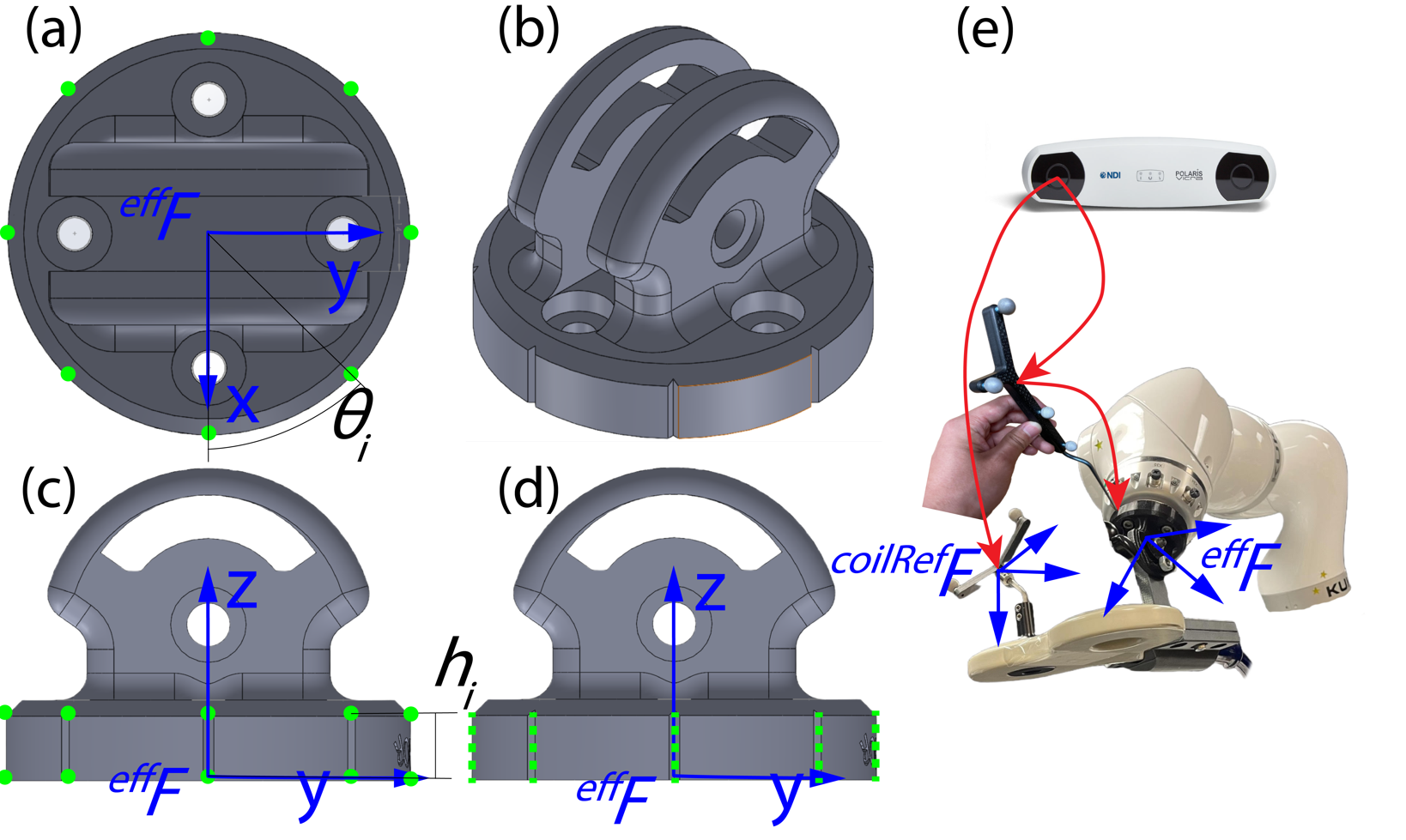}
    \caption{End-effector attachment (TMS coil holder) and landmarks. Fig. (a) is the top view of the end-effector attachment, where green dots represent the landmarks used for paired-point registration. The coordinates of these landmarks with respect to $^{eff}F$ can be determined using $\theta_i$ and $h_i$, as shown in Equation \ref{eq:landmarkcoor}. Fig. (b) provides an isometric view of the end-effector attachment. Fig. (c) shows a side view, illustrating the reference frame $^{eff}F$ represented by blue arrows. The coordinates of the landmarks (green dots) with respect to $^{eff}F$ can be derived by the geometry of the end-effector attachment design. Fig. (d) shows the fitted lines of the grooves. The fitted lines with respect to $^{eff}F$ can also be derived by the geometry of the end-effector attachment design. Fig. (e) shows the digitization of landmarks or grooves.}
    \label{fig:coilholder}
\end{figure}

\begin{figure}[t]
    \centering
    \includegraphics[width=0.4\textwidth]{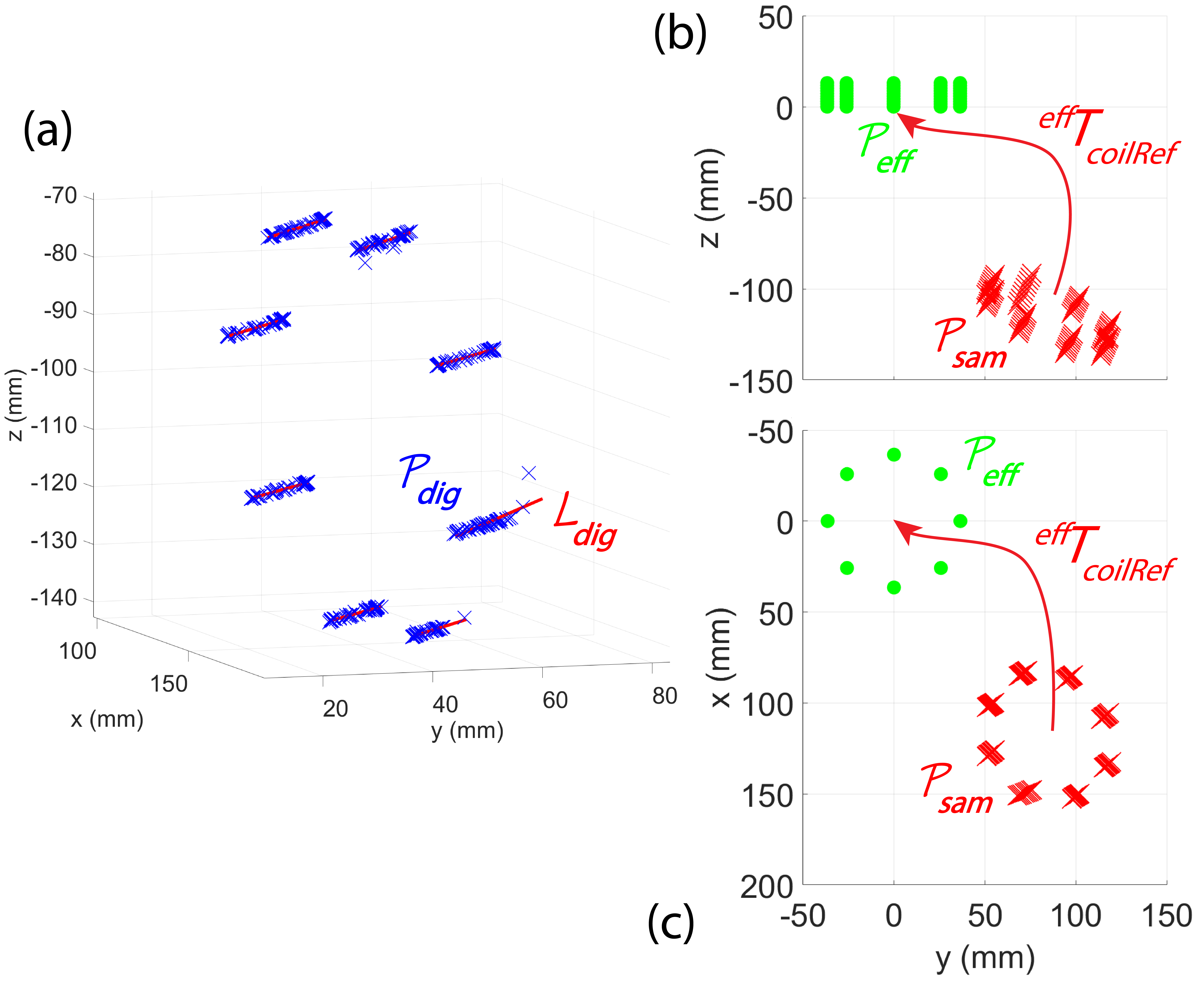}
    \caption{Line fitting using digitized point cloud. (a) shows the digitized point cloud $\mathcal{P}_{dig}$ in blue crosses and the regression lines $\mathcal{L}_{dig}$ in red lines, with respect to the reference frame $^{coilRef}F$. (b) and (c) show the point cloud $\mathcal{P}_{eff}$ (green dots) defined in Fig. \ref{fig:coilholder} and the sampled cloud point $\mathcal{P}_{sam}$ (red crosses) from regression lines $\mathcal{L}_{dig}$. The point clouds are with respect to $^{eff}F$ and $^{coilRef}F$ with overlapped origins (Fig. \ref{fig:coilholder}e). (b) is the side view and (c) is the top view. The result of the paired-point registration is the transformation $^{eff}T_{coilRef}$ (Fig. \ref{fig:alignkinematics}(b)), and the result transformation is denoted in red arrows in (b) and (c).}
    \label{fig:linefit}
\end{figure}

\subsection{Paired-point registration and calibration}\label{sec:pairedpointcal}

We propose to use paired-point registration \cite{Arun1987} to derive the transformation between the TMS coil marker and the end-effector of the robot arm. Instead of the traditional $AX=XB$ or $AX=ZB$ setup, we formulate the transformation as a geometry problem. The key aspect of solving the transformation lies in inferring the center of the end-effector using geometry. Most robot arms are equipped with a circular flange, and we can exploit the property of the circle. Even though here we make the circular shape assumption, the method can be generalized to other geometries. To demonstrate, we provide an example of circular shape in this section, and an asymmetrical example in Section \ref{sec:methodfemoroplasty}. 

By knowing the geometry of the robot arm attachment from Computer Aided Design (CAD), we can establish a coordinate system with its origin situated at the center of the flange. To accomplish this, we design a flange attachment with identifiable landmarks along its rim. The origin of this attachment coincides with the center of the end-effector. Fig. \ref{fig:coilholder}a shows the landmarks calculated from a circular shape with its center aligning with the center of the flange attachment. As such, the coordinates of the landmarks can be derived as:

\begin{equation}\label{eq:landmarkcoor}
    p^{(i)}_{eff}=(cos\theta_i,sin\theta_i,h_i)
\end{equation}

where $\theta_i$ is the angle between $x$ axis (Fig. \ref{fig:coilholder}) and the line connecting center and the landmark $p^{(i)}_{eff}$. $h_i$ is the height from the center to the landmark along $z$ axis. In the case of Fig. \ref{fig:coilholder}a and \ref{fig:coilholder}c, the number of landmarks is 16. The number of landmarks should be larger than or equal to 3 to perform paired-point registration \cite{Arun1987}. It has been reported that the Target Registration Error (TRE) decreases as the number of the landmarks increases \cite{Maurer1997}. We choose 16 to balance the landmark digitization time and accuracy.

With the identifiable landmarks, we can apply paired-point registration to find the transformation between the marker reference $^{coilRef}F$ and the end-effector reference $^{eff}F$. The process of paired-point registration requires digitizing the identifiable landmarks by touching the landmarks using the tip of a rigid body probe. The digitization produces the point cloud $\mathcal{P}_{dig}$, which is a set of points with coordinates based on the marker reference $^{coilRef}F$ on the TMS coil. The solution of paired-point registration is a least-squares fitting optimization, with the minimization criteria defined to be the summation of the Euclidean distances between the original point cloud $\mathcal{P}_{eff}$ (landmarks defined in Fig. \ref{fig:coilholder}) and the transformed point cloud $^{eff}T_{coilRef}(\mathcal{P}_{dig})$:

\begin{equation}
    \Sigma^2=\sum^N_{i=1}||p^{(i)}_{eff}-(Rp^{(i)}_{dig}+t)||^2
\end{equation}

where $i$ is the index of the landmark in the point cloud, $R$ and $t$ are the rotational and translational parts of $^{eff}T_{coilRef}$. The algorithm we use to derive the result of paired-point registration is \cite{Arun1987}.

\subsection{Fiducial localization error optimization}\label{sec:FLEopt}

The error of determining the positions of the landmarks is called Fiducial Localization Error (FLE), defined in \cite{Maurer1997}. In the context of medical image registration, as reported by Maurer \textit{et al.}, FLE exists in both image markers ($FLE_I$) and physical space markers ($FLE_P$). $FLE_I$ corresponds to the error when locating the coordinates of the landmarks on the medical image with blurring and other distortions from the imaging process, further increased by the arbitration of the anatomical landmarks. $FLE_I$ can be trivial in high-fidelity images, and it does not exist with a CAD file of the flange attachment, because the dimensions are easily measurable. $FLE_P$ is more significant in our case, corresponding to the errors during the digitization process by touching physical landmarks using a tracked probe, as shown in Fig. \ref{fig:coilholder}e. Thus, FLE will refer only to $FLE_P$ in this work.

To optimize the digitization process and reduce the FLE, we use fitted lines instead of landmarks to obtain the digitized point cloud. We design the flange attachment to have grooves to replace landmarks, shown in Fig. \ref{fig:coilholder}. Each groove can be digitized as a linear point cloud. A linear point cloud regresses to a straight line, from which we sample another point cloud, $\mathcal{P}_{sam}$. The FLE is expected to reduce because the digitized points of a line are considered to be Gaussian around the true line. This assumption is generally valid if the sensor noise is white noise but it may also be affected by individual differences in performance by operators. The lack of ground truth makes validating reduced FLE difficult, but we design an empirical method to estimate the FLE reduction in Section \ref{sec:resultslineregress}.

The regression method used here can be formulated as an optimization problem in Equation \ref{eq:linefit}:

\begin{equation}\label{eq:linefit}
    \min_{\beta} \sum_{i=0}^N || A(\beta,p_{dig}^{(i)}) - p_{dig}^{(i)}||^2
\end{equation}

where $(i)$ is the $i$th point, $p_{dig}$ is a digitized point, and $A$ represents the projection of the given data point on the fitted line. The goal of the optimization is to find the line parameter $\beta$ that minimizes the sum of the L2-norm between the data points and the fitted line. The results presented in Section \ref{sec:resultslineregress} use SVD to solve the optimization problem \cite{mandel1982use}. 

\subsection{Application on robot-assisted femoroplasty}\label{sec:methodfemoroplasty}

To examine the versatility of GBEC, we also used an asymmetrical set of landmarks to robot-assisted femoroplasty. Because the landmarks are asymmetrical, the same straight line regression in the calibration for TMS does not apply. However, if FLE reduction is needed in an application, straight lines can be identified in the CAD modeling. For example, straight edges of the tool. Grooves can also be manufactured after designing the tool. The tool does not need to be a circular shape, as long as the origin of the tool coincides with the center of the end-effector of the robot.

\begin{figure}[t]
    \centering
    \includegraphics[width=0.28\textwidth]{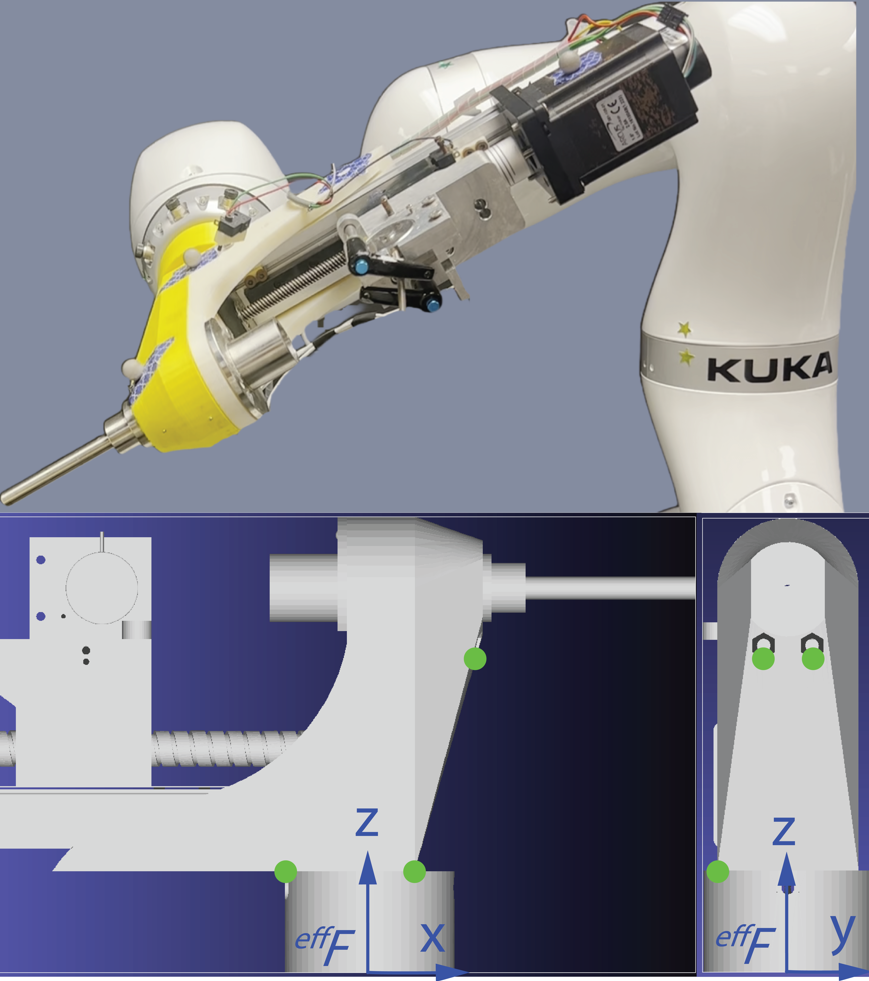}
    \caption{The setup of RDID on the robot and the landmarks used for GBEC. CAD model is shown in front (partially) and side views, with the landmarks (green dots) defined in the reference frame $^{eff}F$ (blue arrows).}
    \label{fig:femoroplasty}
\end{figure}

\begin{figure}[t]
    \centering
    \includegraphics[width=0.45\textwidth]{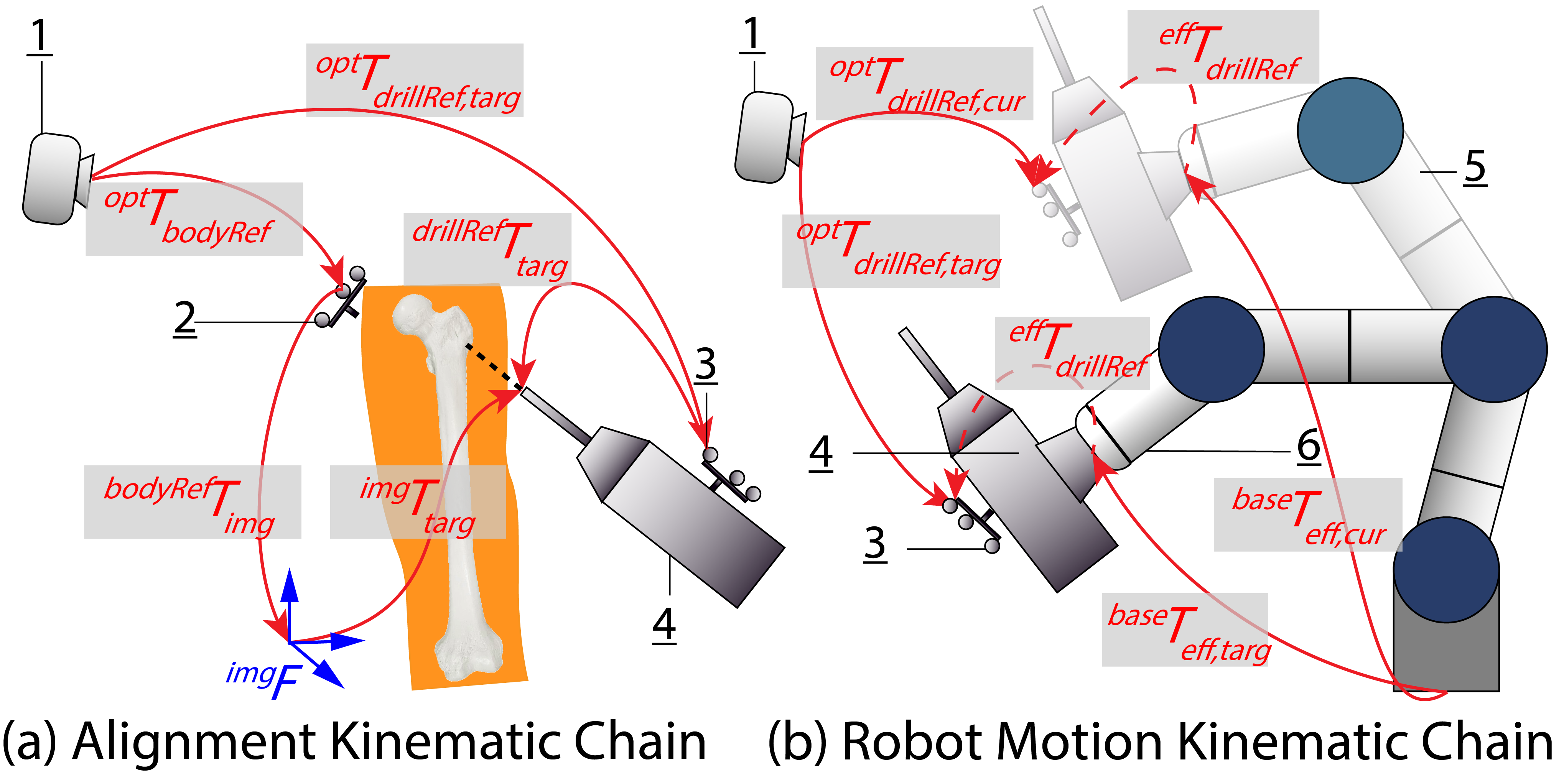}
    \caption{The kinematics for Robot-Assisted Femoroplasty. The numbers \underline{1}, \underline{5}, and \underline{6} are the same as in Fig. \ref{fig:alignkinematics}. Other numbers are: \underline{2} RBM affixed to the patient, \underline{3} RBM affixed to RDID, and \underline{4} RDID. The hand-eye calibration result is the dashed line arrow.}
    \label{fig:femuralignkinematics}
\end{figure}

In femoroplasty, minimally-invasive osteoporotic hip augmentation is a procedure to inject bone cement into the femoral neck and head, which reduces the risk of hip fracture. The tool of the robotic system is a custom-designed Robotic Drilling and Injection Device (RDID, Fig. \ref{fig:femoroplasty}). Multiple previous works have investigated surgical trajectory planning and execution \cite{basafa2015subject, farvardin2021biomechanically, otake2010image, horbach2020biomechanical}, and our group has developed robotic methods to improve drill accuracy, reduce cementation errors, and enhance safety and efficiency \cite{bakhtiarinejad2023surgical}. Shown in Fig. \ref{fig:femuralignkinematics}, aiming to align the drill and injector to a planned trajectory on the patient's femur, robot-assisted femoroplasty shares the same kinematics as robot-assisted TMS. The procedure also requires the transformation between the end-effector and the marker attached to RDID, to complete the kinematics chain. Fig. \ref{fig:femoroplasty} shows the RDID used in the experiment and the definition of the 4 asymmetrical landmarks.

\section{Results and Discussion}\label{sec:results}

We examine the accuracy of the calibration for robot-assisted TMS by decomposing the sources of errors and investigating them from lower level to higher level along the propagation of the errors. At the lowest level, we obtain data from the optical tracking camera, which is considered to be accurate. At a higher level, we examine line regression errors, FLEs, registration residuals, and repeatability by repeating GBEC and AXXB for robot-assisted TMS. We also investigate the independency of our approach to workspace, and show alignment errors. The results are compared to previous works whenever applicable. For robot-assisted femoroplasty, we repeat GBEC and AXXB to show registration residuals, repeatability, workspace independency, and overall errors.

\subsection{Line regression errors and FLE reduction (TMS)} \label{sec:resultslineregress}

\begin{figure}[t]
    \centering
    \includegraphics[width=0.4\textwidth]{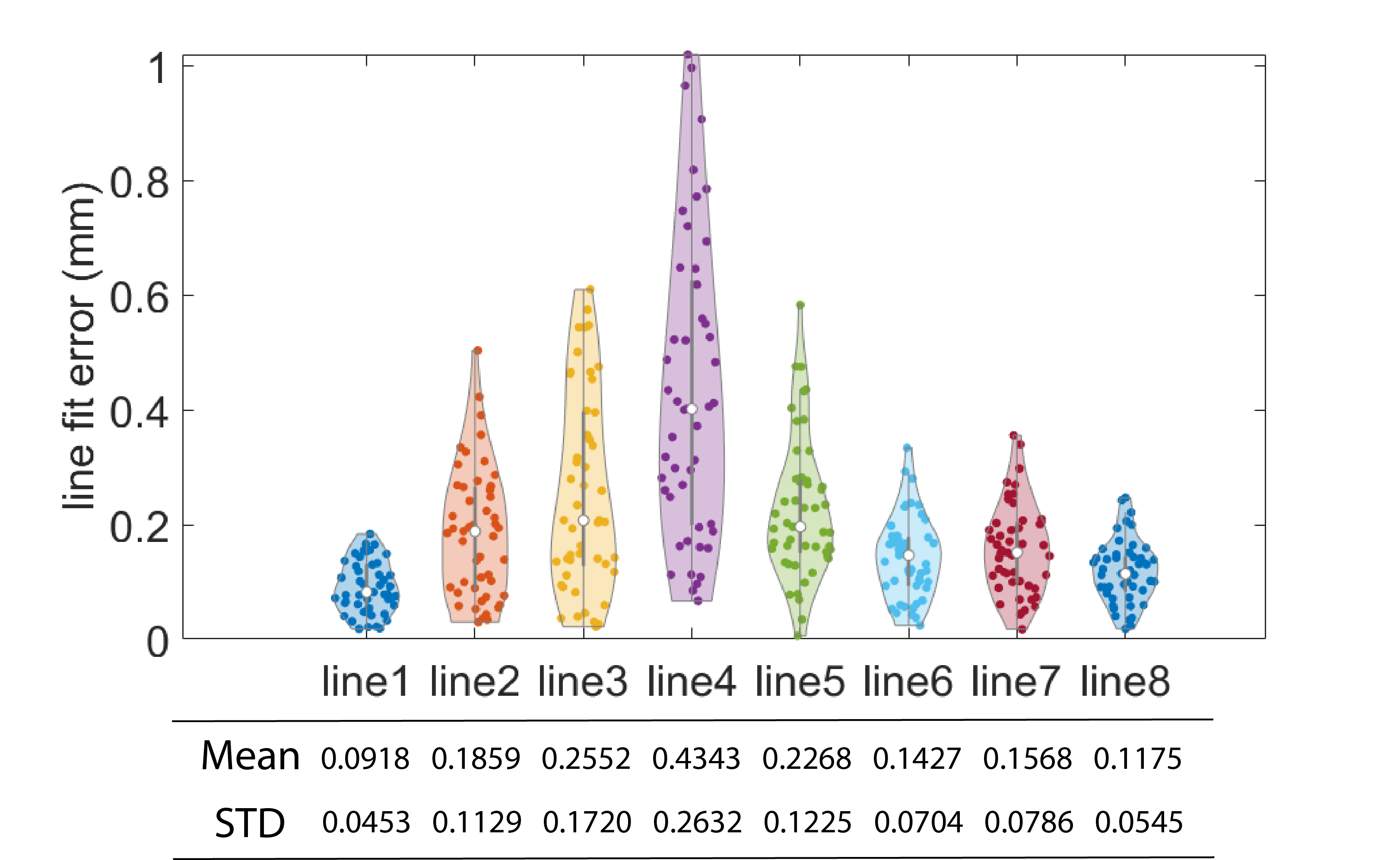}
    \caption{Line regression errors in GBEC used for robot-assisted TMS. Lines are shown separately because the difficulty of manually digitizing each line can be different, which contributes to different errors.}
    \label{fig:fitlineerr}
\end{figure}

\begin{figure}[t]
    \centering
    \includegraphics[width=0.4\textwidth]{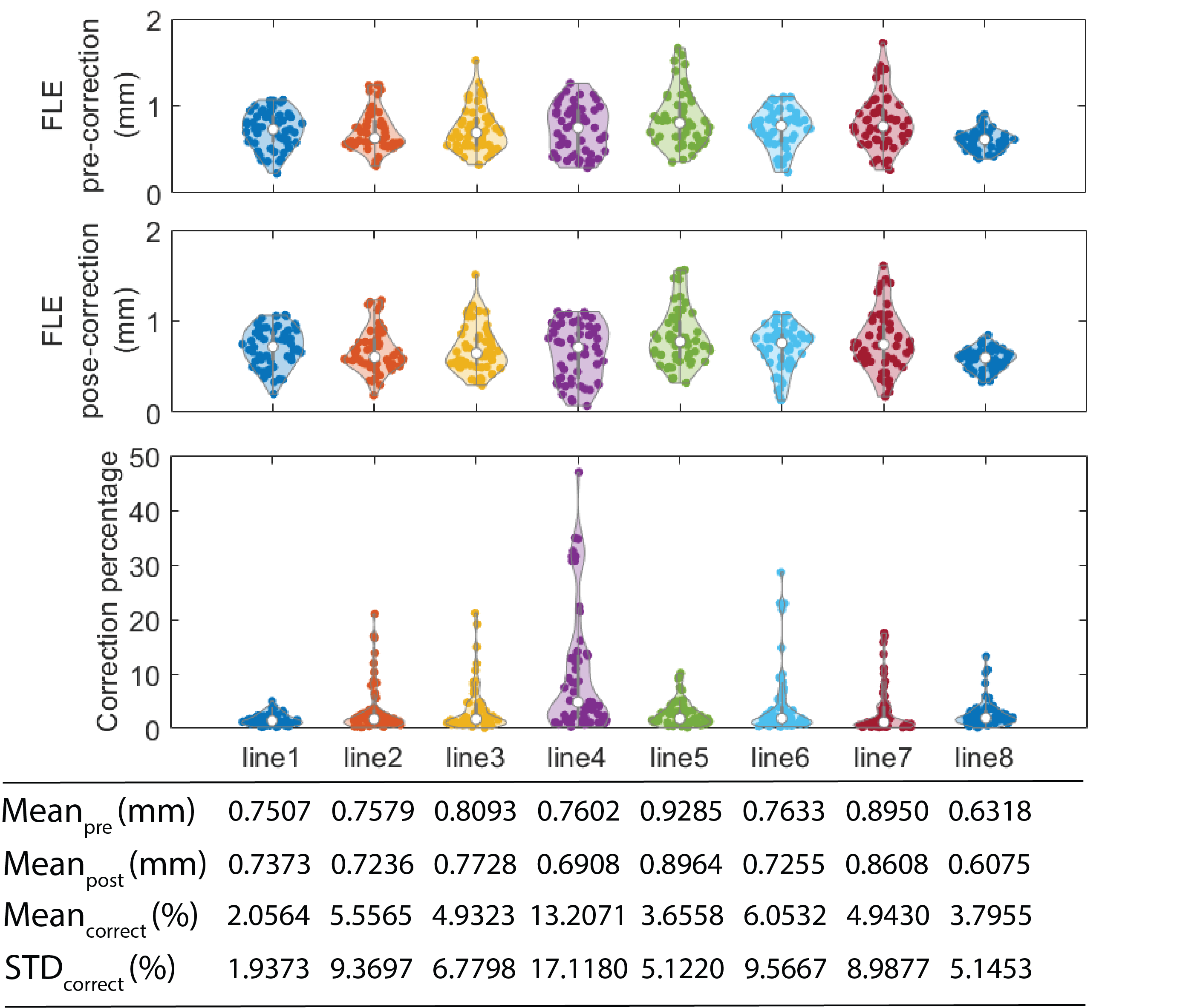}
    \caption{FLE and its correction by line fitting for the hand-eye calibration for robot-assisted TMS.}
    \label{fig:FLE}
\end{figure}

For robot-assisted TMS, each in the 57 calibration trials has 8 lines as shown in Fig. \ref{fig:coilholder}a-d and \ref{fig:linefit}. Each line is digitized manually to obtain $\mathcal{P}_{dig}^{(i)}$ containing 52 digitized points and the fitted $\mathcal{L}_{dig}^{(i)}$, shown in Fig. \ref{fig:linefit}a where $i$ refers $i$th line. The line regression error of a digitized point is calculated by the shortest distance from the point in the digitized point cloud $\mathcal{P}_{dig}$ to the fitted line $\mathcal{L}_{dig}$. An example of the line fitting of a calibration trial is shown in Fig. \ref{fig:linefit}a, where a digitized point cloud of grooves is in blue, denoted as $\mathcal{P}_{dig}$, and the fitted lines are in red, denoted as $\mathcal{L}_{dig}$. 

Fig. \ref{fig:fitlineerr} shows the regression errors, where each solid circular point represents the mean error of all the 52 digitized points in a point cloud $\mathcal{P}_{dig}$. Except for line \#4, all lines have a mean fitted error of around 0.2 mm. The reason is that line \#4 is behind the TMS coil and is facing the opposite way of the optical tracking camera, which causes inaccuracies in digitization. 

To validate FLE reduction, a ground truth is needed by transforming the original point cloud $\mathcal{P}_{eff}$ to the reference frame $^{coilRef}F$, for deriving $\hat{\mathcal{P}}_{coilRef}$. Here, we used the calibration result from each trial, transform the defined lines (Fig. \ref{fig:coilholder}d) from $^{eff}F$ to $^{coilRef}F$, and used the transformed lines as the ground truth. 

Pre-correction FLE is the distance from the ground truth to $\mathcal{P}_{dig}$, whilst post-correction FLE is the distances from the ground truth to the projection of $\mathcal{P}_{dig}$ on $\mathcal{L}_{dig}$. Note the FLE shown here is an estimate using a ground truth that causes a balance of FLEs among the lines. Still, we can observe the largest recovery of accuracy on line 4 using FLE optimization. Our results demonstrate a reduction of FLE in all lines in 57 calibration trials by the proposed line-fitting method, as shown in Fig. \ref{fig:FLE}. The mean of pre-correction FLE all reduced, and the mean percentage of correction is up to 13\% (three outliers are eliminated from each line, so the reduction can be higher). The method is particularly effective in lines that are not well-digitized. For example, line 4 has the highest line regression residual, shown in Fig. \ref{fig:fitlineerr}, and the proposed method reduces its FLE by the highest percentage, shown in Fig. \ref{fig:FLE}. Our results show that the linear regression and line localization do not contribute to the propagation of sensor errors, because the result line regression errors in Fig. \ref{fig:fitlineerr} and FLEs in Fig. \ref{fig:FLE} are below 1 mm, which is at a similar level to the errors from the sensor.

\subsection{Landmark residuals (TMS and femoroplasty)}\label{sec:resultsregres}

\begin{table}
\caption{Residual of the paired-point registration for 39 GBEC trials for robot-assisted femoroplasty }
\label{tab:residual}
\newcolumntype{Y}{>{\centering\arraybackslash}X}
\begin{tabularx}{0.49\textwidth}{cYYYY}
\toprule
Landmark& 1&2&3&4\\
\midrule
STD (mm)&0.2744&0.3067&0.3772&0.4005\\
Mean (mm)&0.8768&0.9059&0.8674&0.8375\\
\bottomrule
\end{tabularx}
\end{table}

\begin{figure}[t]
    \centering
    \includegraphics[width=0.4\textwidth]{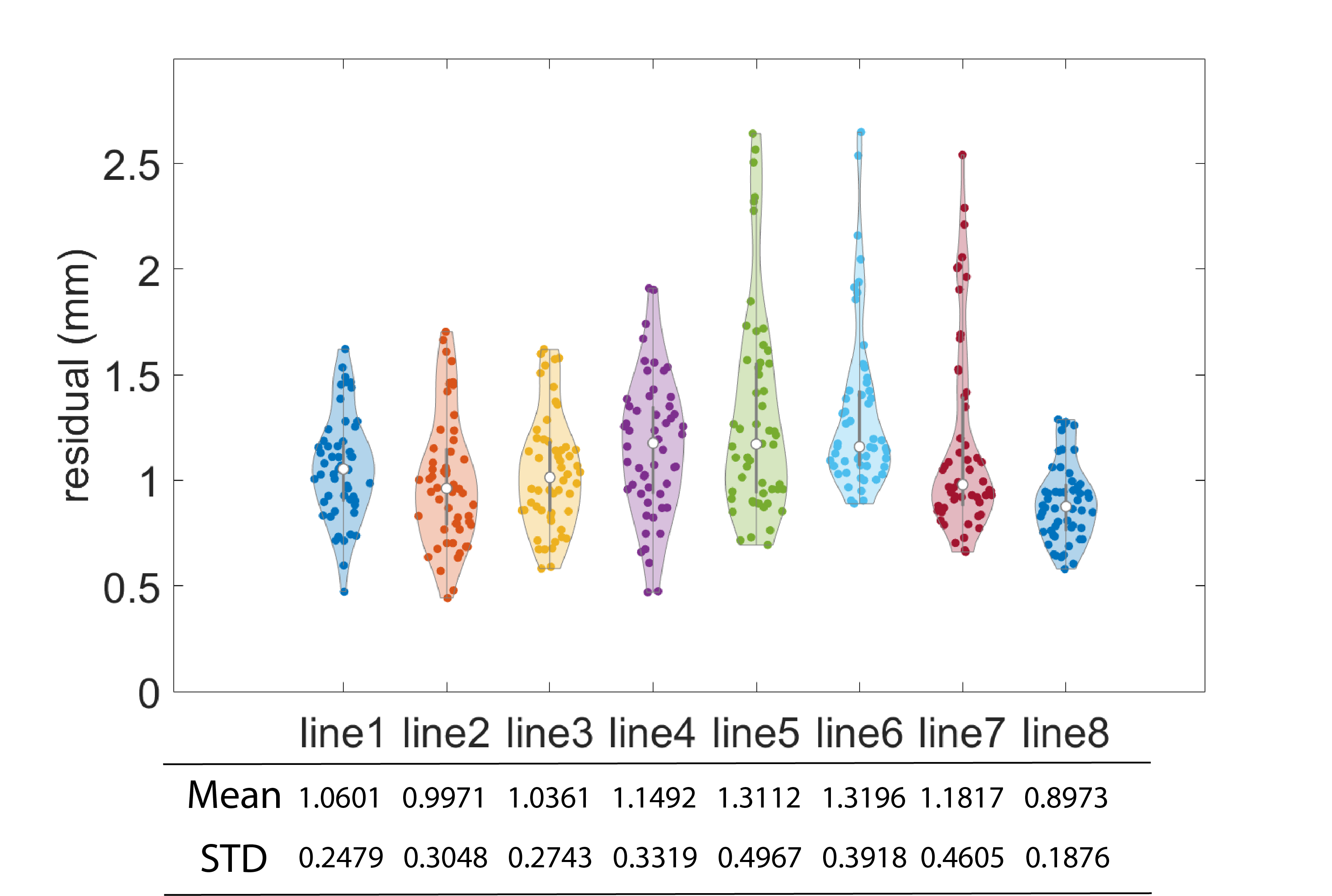}
    \caption{The residual of the paired-point registration for 57 calibration trials for the hand-eye calibration for robot-assisted TMS. Each column is the distribution of the residuals on a line. Three outliers are removed for each line. }
    \label{fig:regresidual}
\end{figure}

Landmark residual is the discrepancy between the transformed points in one coordinate system and their corresponding points in another coordinate system during a registration process. We use Euclidean distances to quantify this error:

\begin{equation}
    \mathcal{E}=\mathcal{P}_{eff}-^{eff}T_{coilRef}(\mathcal{P}_{sam})
\end{equation}

For robot-assisted TMS, as Fig. \ref{fig:regresidual} shows, the paired-point registration for 57 calibrations on each line has a mean of about 1 mm. The mean error of all lines is $1.12\pm0.38$ mm (mean $\pm$ std). In Fig. \ref{fig:regresidual}, each residual is calculated as the mean residual of 10 points on a line, with $x,y,z$ directions averaged. As such, each residual is interpreted to be the mean residual of a line in a calibration trial. Our results show small registration residuals, proving the data collection in GBEC and the result transformation to be valid. For robot-assisted femoroplasty, the landmarks used for the calibration have a residual of $0.87\pm0.34$ mm, in 39 GBEC trials (Table \ref{tab:residual}).

\subsection{Calibration repeatability and workspace independency (TMS and femoroplasty)}\label{sec:calrep}

\begin{figure}[t]
    \centering
    \includegraphics[width=0.32\textwidth]{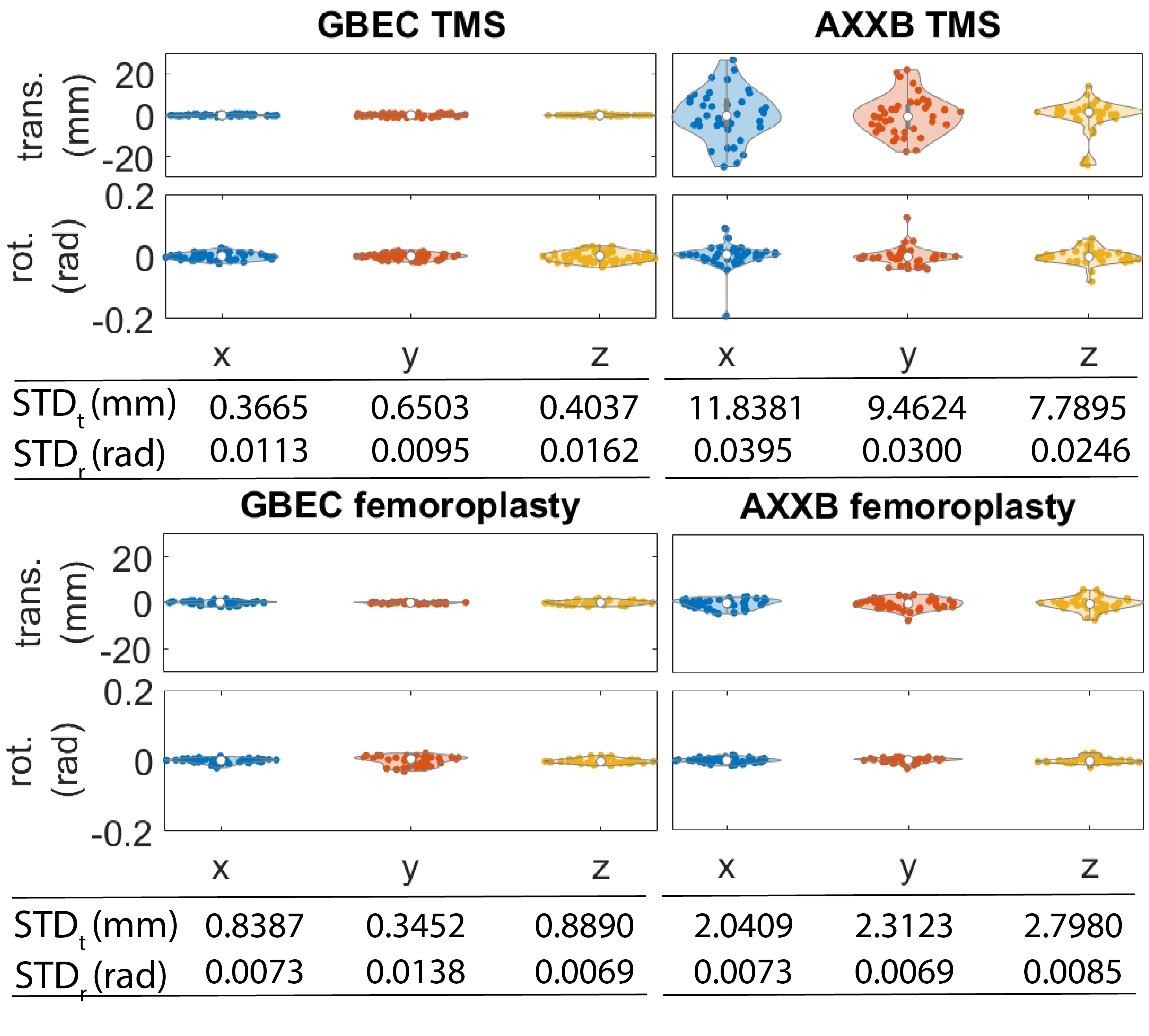}
    \caption{The repeated calibration results. 57 GBEC and 39 AXXB have been performed for robot-assisted TMS, and 39 GBEC and 39 AXXB have been performed for robot-assisted femoroplasty. GBEC outperforms AXXB in terms of the standard deviation of the results of the calibration.}
    \label{fig:calrepeat}
\end{figure}

\begin{figure}[t]
    \centering
    \includegraphics[width=0.35\textwidth]{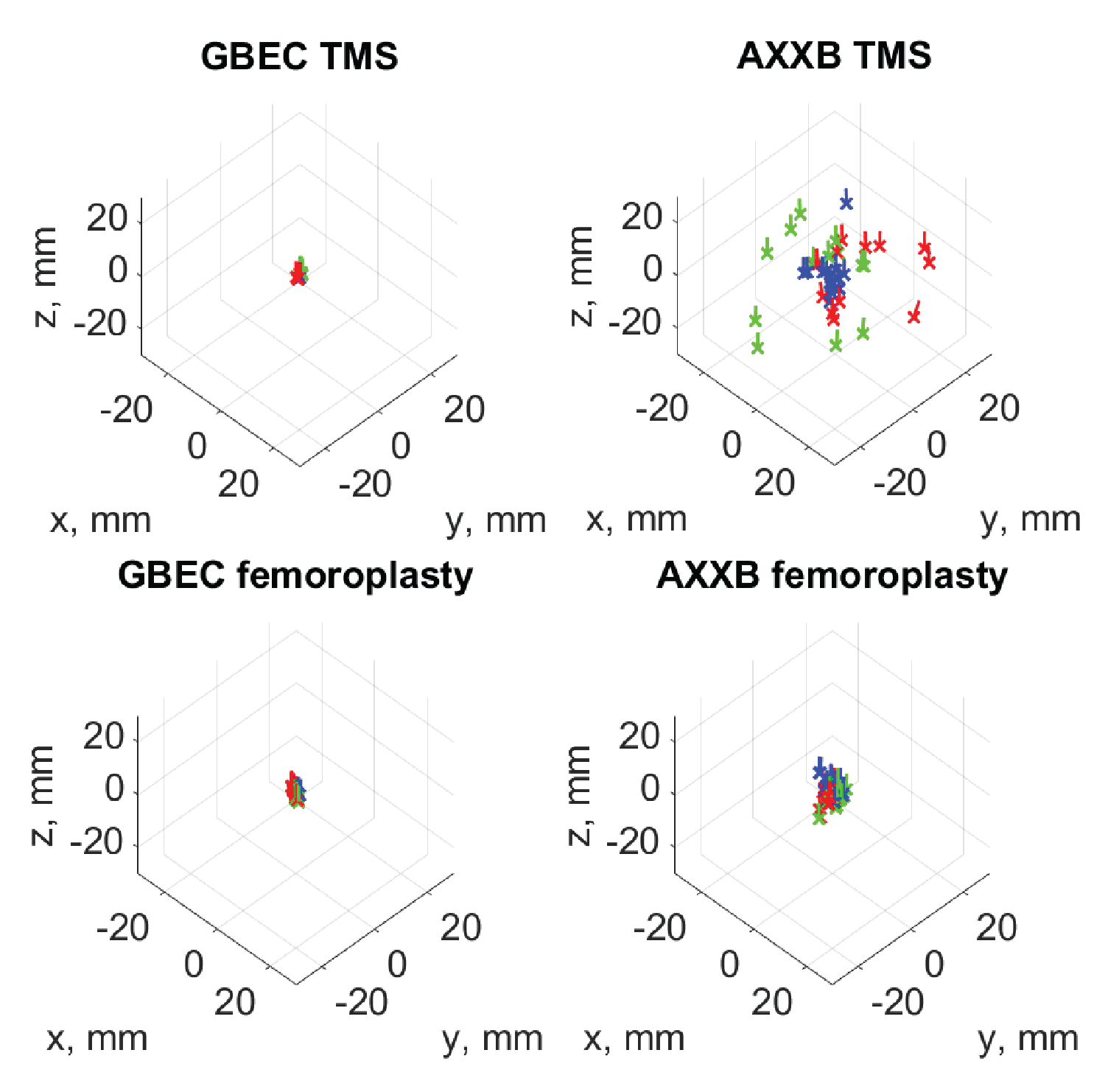}
    \caption{The transformations derived by GBEC and AXXB at 3 different robot workspaces. The transformations are visualized as vectors, using the position to represent translation and direction to represent rotation. The compactness of the vectors represents repeatability.}
    \label{fig:workspace}
\end{figure}

The true calibration result depends only on the geometry of the flange attachment, so repeated calibrations should derive the same result. However, for robot-assisted TMS, Richter \textit{et al.} \cite{richter2011robust} reported a translation variation of 2.94mm, 6.43mm, and 0.75mm for QR24\cite{ernst2012non}, Tsai-Lenz \cite{tsai1989new}, and online calibration \cite{richter2011robust}, respectively. Whether these variations from the three different methods were standard deviation or the range of the results is not given in the article. Our method produces results (Fig. \ref{fig:calrepeat}) with a standard deviation of 0.37, 0.65, 0.40 mm along x, y, z axes (TMS) and 0.84, 0.35, 0.89mm (femoroplasty), showing high repeatability of GBEC. Additionally, the calibration results are superior than AXXB for both TMS and femoroplasty setups.

The result transformations from GBEC do not rely on robot joints. To provide evidence, we conducted experiments by repeating GBEC and AXXB in 3 distinct robot workspaces. At each workspace, we performed 8 (GBEC) and 13 (AXXB) calibrations and obtained the resulting transformations. In each calibration trial using AXXB formulation, we ensure: (1) Collected data contains at least 50 different robot poses, (2) The workspace of the end-effector, from which the robot data is collected, is a spherical space with a diameter of 0.4 meters, (3) The orientation of the robot end-effector is generated randomly in a range of 15 degrees in any axis, allowing robot joints to have large changes between different poses. Fig. \ref{fig:workspace} visualizes the transformations as vectors, and the transformations at the same workspace are plotted in the same color. The cluster in GBEC is tightly distributed within a range of 2mm in all axes, and significant separation of the vectors on robot poses is not observed. This is also consistent with the repeatability of the transformations shown in Fig. \ref{fig:calrepeat}.

\subsection{Alignment accuracy (TMS and femoroplasty)}\label{sec:endtoend}

We performed alignments using the calibrated tools for both robot-assisted TMS and femoroplasty on an integrated medical robotic system \cite{liu2023toward}, and investigate the error between the target pose and measured pose of the tools after alignment. A laser-scanned foam head and a femur phantom with a CT scan were used in the experiments and the registration residual for both were less than 1.5mm. Table \ref{tab:align} shows the mean and the standard deviation of the translational errors and rotational errors in 12 alignments for each application. To compare robot-assisted TMS alignment errors with previous works, we have achieved accuracy of: Translation, mean$\pm$std: $0.12\pm0.08 mm,  0.26\pm0.17 mm, 0.37\pm0.40 mm$; rotation, mean$\pm$std: $0.08\pm0.04^{\circ} , 0.11\pm0.07^{\circ} , 0.08\pm0.05^{\circ} $. The aforementioned findings indicate a higher alignment accuracy compared to Richter \textit{et al} \cite{richter2011robust}. In their study, they achieved Euclidean distances, accounting for all axes, of 1.80mm, 7.12mm, and 2.21mm for QR24 \cite{ernst2012non}, Tsai-Lenz \cite{tsai1989new}, and the online calibration method \cite{richter2011robust}, respectively. Robot-assisted femoroplasty achieve an error around 0.3 mm.

\begin{table}
\caption{Tool alignment errors using GBEC, represented by the distance between the planned pose and the measure pose after each alignment.}
\label{tab:align}
\newcolumntype{Y}{>{\centering\arraybackslash}X}
\begin{tabularx}{0.49\textwidth}{cYYYYYY}
\toprule
Axis& x (mm)&y (mm)&z (mm)&rx (deg)&ry (deg)&rz (deg)\\
\midrule
STD (TMS)&0.0775&0.1701&0.3994&0.0443&0.0735&0.0507\\
Mean (TMS)&0.1171&0.2646&0.3694&0.0833&0.1061&0.0787\\
\midrule
STD (Fem.)&0.2200&0.2948&0.2279&0.0843&0.0534&0.0433\\
Mean (Fem.)&0.3072&0.3788&0.2158&0.1125&0.0571&0.0432\\
\bottomrule
\end{tabularx}
\end{table}

% \begin{figure}[ht]
%     \centering
%     \includegraphics[width=0.4\textwidth]{figs/placeerror.png}
%     \caption{Tool alignment errors, represented by the distance between the planned pose and the measure pose after each alignment.}
%     \label{fig:placeerror}
% \end{figure}

\subsection{Camera-in-hand setup}\label{sec:future}

% In the preceding sections, we first outline the motivations behind employing a geometry-based hand-eye calibration approach. Subsequently, we introduce the kinematics involved and derive the transformation to be computed. We then elucidate the geometry methodology to determine the transformation between the end-effector and the attached marker. Moreover, we conduct a comprehensive analysis of the errors, beginning from lower-level to higher-level considerations. Additionally, we touch upon the extension of the method to accommodate asymmetric landmarks. 

\begin{figure}[ht]
    \centering
    \includegraphics[width=0.45\textwidth]{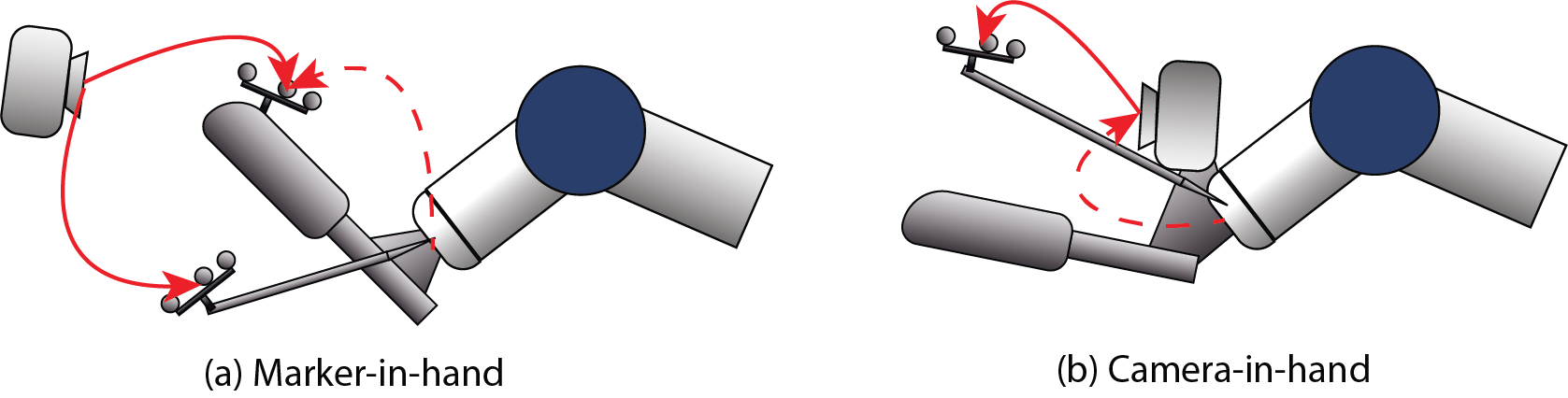}
    \caption{Generalization of GBEC in ``marker-in-hand'' and ``camera-in-hand'' setups. The figure contains a tracked probe to digitize landmarks on the end-effector.}
    \label{fig:generalization}
\end{figure}

We described the methodology and provided data analysis on ``marker-in-hand'' setup. However, ``camera-in-hand'' setup can also use the geometric features to conduct hand-eye calibration, thus GBEC can be applied to the two most common setups for hand-eye calibration. Fig. \ref{fig:generalization} demonstrates ``marker-in-hand'' and ``camera-in-hand'' setups, along with the application of GBEC using a probe. In a ``marker-in-hand'' setup, the goal of hand-eye calibration is to establish the transformation between the end-effector and the marker. On the other hand, in the ``camera-in-hand'' setup, the objective is to determine the transformation between the end-effector and the origin of the camera (sensor). To this end, we summarize 3 conditions for a successful GBEC use case: (1) The end-effector attachment has identifiable and digitizable landmarks, (2) the end-effector (or robot flange) center can align with a point whose relative location is geometrically derivable by the landmarks on the end-effector, and (3) the landmarks on the end-effector attachment need to be visible by the sensor to be calibrated. Visibility can mean direct accessibility or indirect accessibility by a kinematic chain.

\section{Conclusion}

In this work, we introduce GBEC, a hand-eye calibration method that leverages the geometric features of the robot flange and its attached marker. Our experimental results demonstrate that GBEC is accurate, repeatable, independent of workspace or robot pose, and only rely on the geometry of the end-effector attachment. We extend the method to accommodate both symmetrically and asymmetrically placed landmarks on the end-effector, making it applicable in various applications such as robot-assisted TMS and femoroplasty. In particular, our application of robot-assisted TMS shows superior repeatability and tool alignment accuracy than the current literature. Furthermore, we broaden the scope and generalize the setup from ``marker-in-hand'' to ``camera-in-hand'', addressing the most common cases to use hand-eye calibration. 

\bibliographystyle{IEEEtran}
\bibliography{bib} % bibliography data in bib.bib

\end{document}